\documentclass[letter, 10pt, conference]{ieeeconf}
\IEEEoverridecommandlockouts                              
\usepackage{graphicx}
\usepackage{amsmath}
\usepackage{amsfonts}
\usepackage{subcaption}
\usepackage{booktabs}
\usepackage{multirow}
\usepackage{url}
\usepackage{hyperref}
\usepackage{cleveref}
\usepackage{comment}
\usepackage[acronyms]{glossaries}
\usepackage{xcolor}

\crefname{section}{Sec.}{Secs.}
\crefname{table}{Tab.}{Tabs.}
\crefname{figure}{Fig.}{Figs.}
\crefname{equation}{Eq.}{Eqs.}

\title{\LARGE \bf
RGB-only Active 3D Scene Graph Generation for Indoor Mobile Robots
}

\author{Giorgia Modi$^{12*}$, Davide Buoso$^{2}$, Giuseppe Averta$^{2}$, Daniele De Martini$^{1}$ \\ 
$^{1}$Mobile Robotics Group (MRG), University of Oxford, UK \\
$^{2}$Visual and Multimodal Applied Learning Lab (VANDAL), Politecnico di Torino, Italy
\thanks{*Correspondent author: \texttt{giorgiamodi@gmail.com}}
\thanks{This work was supported by the EPSRC Impact Acceleration Account (IAA), and the UKRI InnovateUK project `A general-purpose configuration management system for highly configurable systems validated in automated logistics'. 
}
}

\begin{document}

\maketitle

\begin{abstract}
Current approaches to 3D scene graph generation rely on dedicated depth sensors, such as LiDAR or RGB-D cameras, for metric 3D reconstruction. This limits deployment to specialized robotic platforms and excludes settings where only RGB cameras are available, such as fixed external infrastructure. Existing pipelines also typically operate on passively collected observation trajectories, rather than selecting viewpoints based on the partially built scene representation, and therefore fail to effectively exploit the semantic and spatial information encoded within the graph during exploration. This paper presents a fully visual framework for the active, incremental construction of 3D scene graphs from RGB input only, addressing both limitations. The proposed approach unifies perception and planning around a shared structured representation that captures object semantics, 3D geometry, relational context, and information from multiple viewpoints. Because the framework is hardware-agnostic and relies only on RGB observations, it can incorporate inputs from both onboard robot cameras and fixed external cameras within the same representation.
Experiments on the Replica dataset show that the RGB-only pipeline achieves F1-score parity with baselines using ground-truth depth. Active exploration experiments on ReplicaCAD further show that semantic-driven viewpoint selection detects more than twice as many objects as a geometric frontier-based baseline under the same exploration budget. Finally, the external-camera setting demonstrates that complementary RGB views can effectively bootstrap the scene graph and improve contextual understanding at no additional exploration cost.
\end{abstract}

\section{Introduction}
Modern robotic systems operating in unstructured human environments—such as homes, hospitals, and offices—depend on rich contextual scene understanding to perceive, decide, and act autonomously~\cite{ni2023deep}. Traditional perception pipelines either build low-level dense metric maps that capture free and occupied space but lack semantic context, or produce object-centric outputs such as bounding boxes and semantic segmentations, which isolate entities effectively yet fail to encode relational structure and distinguish semantically different but visually similar spatial configurations~\cite{xu2017iterative}. Reliable robot autonomy therefore requires structured scene representations that jointly capture which entities are present, where they are located in three-dimensional space, and how they relate to one another.

To address these limitations, 3D scene graphs have emerged as a foundational architecture for environment perception and contextual reasoning~\cite{Li2024survey,bae2023survey3Dscenegraphs}. A 3D scene graph models the environment as a graph whose nodes represent object instances with semantic labels and metric locations, while edges encode pairwise spatial and functional relationships. This compact, queryable representation bridges low-level visual perception and high-level decision-making, supporting language grounding and task planning~\cite{rana2023sayplan}, navigation~\cite{werby2024hierarchical}, and manipulation in partially observed scenes~\cite{jiang2024roboexp}.

Despite their theoretical utility, the deployment of 3D scene graphs is constrained by significant limitations. State-of-the-art frameworks typically require RGB-D cameras or LiDAR to acquire the dense depth required for metric 3D reconstruction \cite{gu2024conceptgraphs, hughes2022hydra, wu2021scenegraphfusion}. This restricts deployment to specialized platforms, excluding scenarios where only commodity RGB cameras are available, such as surveillance infrastructure, low-cost robots, or handheld devices.  Furthermore, existing 3D scene graph pipelines are predominantly passive, relying on pre-recorded observation trajectories rather than actively selecting informative viewpoints for efficient 3D scene graph construction, thus failing to fully exploit the rich information encoded within the graph.

To address these limitations, we present a fully visual perception--action framework for active 3D scene graph construction from RGB images only, combining MapAnything-based feed-forward 3D reconstruction~\cite{keetha2025mapanything}, ConceptGraphs open-vocabulary semantic mapping~\cite{gu2024conceptgraphs}, and semantic exploration~\cite{tang2025active} within a single pipeline. Our contributions are:

\begin{enumerate}
    \item an RGB-only pipeline for 3D scene graph generation;
    \item the integration of this representation into an active semantic exploration loop for next-best-view selection;
    \item an evaluation of fixed external RGB cameras as complementary observations for scene graph bootstrapping and improvement.
\end{enumerate}

\section{Related Work}
\label{sec:related_work}

\begin{figure*}[t]
    \centering
    \includegraphics[width=0.95\textwidth]{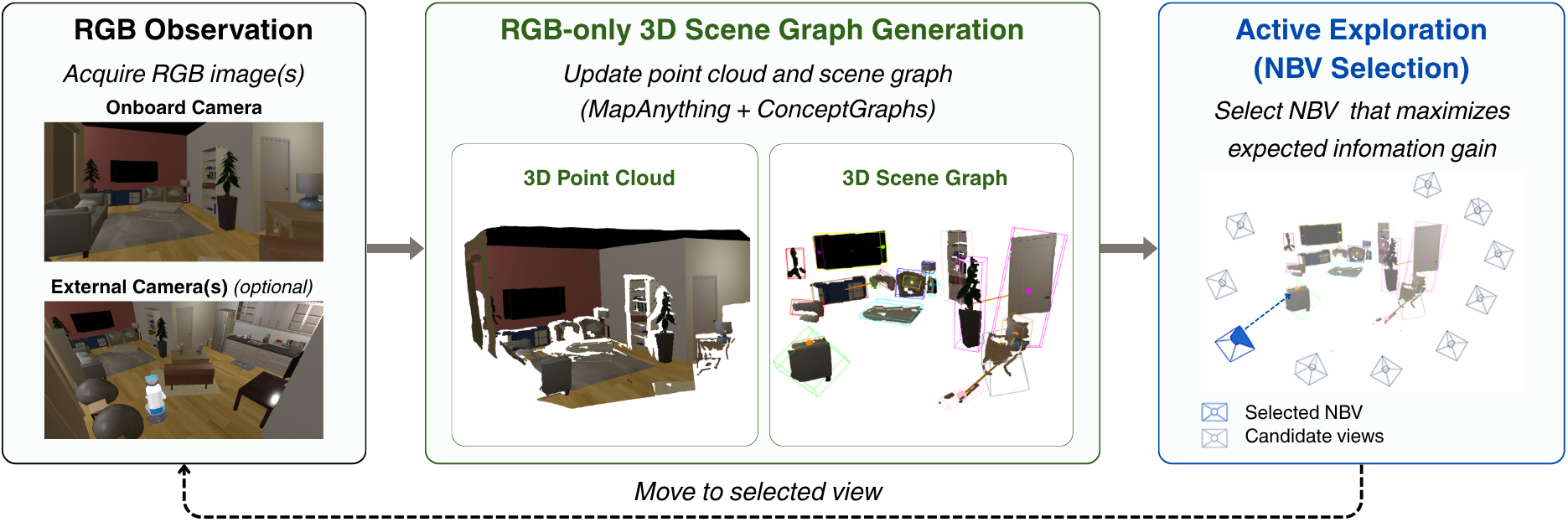}
    \caption{RGB-only active 3D scene graph generation loop. RGB images from the onboard camera (and, optionally, from fixed external cameras) feed the RGB-only pipeline, incrementally updating the point cloud and scene graph. The active module uses the current scene graph to select the NBV. The robot moves to the selected view and the loop repeats.}
    \label{fig:overview}
\end{figure*}

\textbf{3D Scene Graphs (3DSGs).}
Scene graphs were introduced as structured representations coupling object semantics with relational context, addressing ambiguities in purely object-centric systems \cite{xu2017iterative}. Extending this abstraction to 3D grounds entities and relationships in metric space, enabling navigation, manipulation, and language-guided planning~\cite{catalano20253D}. Representative systems such as Hydra \cite{hughes2022hydra} and SceneGraphFusion \cite{wu2021scenegraphfusion} build hierarchical representations from SLAM outputs, relying on posed RGB-D data, but operate on passive trajectories and use closed-vocabulary semantics. Open-vocabulary methods, notably ConceptGraphs \cite{gu2024conceptgraphs}, overcome vocabulary limits using foundation models for flexible language-grounded representations, yet still assume RGB-D inputs. Active scene graph construction remains underexplored: most approaches operate on fixed trajectories without influencing viewpoint selection \cite{catalano20253D}. Exceptions include the embodied framework of Li et al.~\cite{li2022embodiedssgg} and Active Semantic Perception (ASP) \cite{tang2025active}, which uses Large Language Model (LLM)-sampled scene graph completions to guide exploration, but still relies on depth observations and known camera poses.

\textbf{Active Perception.}
Active perception—controlling sensing to maximize information gain—has long been studied via Next-Best-View (NBV) planning \cite{bajcsy2018revisiting}. Geometric frontier-based methods \cite{yamauchi1997frontier} and density-based approaches such as Surface Edge Explorer (SEE) \cite{border2024surface} efficiently expand coverage but ignore semantics. Recent work incorporates semantic reasoning: Ding et al.~\cite{ding2025sea} use semantic scene completion to guide exploration toward uncertain regions, while Chen et al.~\cite{chen2025understanding} combine semantic and geometric uncertainty via 3D Gaussian Splatting.  ASP \cite{tang2025active} further enables LLM-based reasoning over scene graphs, turning them into active exploration tools. However, prior work on ASP has not considered settings with external camera inputs, nor RGB-only perception.

\textbf{Multi-View Perception via External Cameras}
While robotic perception is typically egocentric, fixed external cameras provide complementary global context. Prior work shows their value for visual servoing \cite{robinson2026robot} and VLM-guided navigation \cite{buoso2025select2plan}, but their integration into 3DSGs remains unexplored. 

\section{Methodology}
\label{sec:methodology}

\subsection{System Overview}
The system operates as an incremental perception--action loop (Fig.~\ref{fig:overview}).
At each step $t$, the robot captures an RGB image $I_t$ from its current viewpoint $v_t \in \mathcal{V}$, where $\mathcal{V}$ is the set of navigable viewpoints in the environment.
The RGB-only 3DSG generation pipeline processes $I_t$ to update the reconstructed point cloud $\mathcal{P}_t$ and the scene graph $\mathcal{G}_t = (\mathcal{O}_t, \mathcal{E}_t)$, where $\mathcal{O}_t$ is the set of detected object nodes and $\mathcal{E}_t$ the set of spatial relationship edges between them.
The active exploration module then selects the Next-Best-View (NBV) $v_{t+1}$ that maximizes expected information gain.
The robot navigates to $v_{t+1}$ and the loop repeats.
Fixed external cameras inject additional RGB frames into the pipeline, contributing complementary viewpoints without robot motion.

\subsection{RGB-only 3DSG generation}
\textbf{Depths and Poses Estimation.} 
The pipeline leverages MapAnything~\cite{keetha2025mapanything} to infer scene geometry from a set of RGB images. Given $N$ inputs, the model predicts in a single forward pass a \emph{factored} representation of per-view pixel ray directions $R_i$, up-to-scale depth maps $\tilde{D}_i$, poses $\tilde{P}_i$ expressed in the reference frame of the first image, and a global metric scale factor $m \in \mathbb{R}$ shared across views.
These factored outputs are mathematically composed to back-project 2D pixels into a metrically scaled 3D point cloud. 

Since MapAnything predicts all camera poses in the reference frame of the first input image, recovering absolute poses only requires fixing this reference: knowing the pose of the first input image (e.g., an external camera or the robot’s initial viewpoint) is sufficient to express all poses in a consistent global frame.

\textbf{Open-Vocabulary Scene Graph Construction.}
RGB images, together with estimated depths and poses, are passed to ConceptGraphs~\cite{gu2024conceptgraphs}: SAM~\cite{kirillov2023segment} extracts class-agnostic instance masks, CLIP~\cite{radford2021clip} embeds each region into a semantic descriptor, the masks are projected into 3D using the MapAnything-derived depth and poses, and multi-view association incrementally merges detections into 3-D object nodes $o_j \in \mathcal{O}_t$. While the original ConceptGraphs pipeline queries proprietary LLMs to infer relationships between objects~\cite{gu2024conceptgraphs}, here spatial edges are instead derived through a deterministic, geometry-only procedure over oriented 3-D bounding boxes of the object nodes: for each ordered object pair, a fixed-priority set of predicates is evaluated - \textit{on top of / supported by} (vertical contact with footprint overlap), \textit{under / over} (vertical adjacency), \textit{inside} (volumetric containment), and \textit{next to} (horizontal proximity at similar height) - and at most one edge per ordered pair is assigned, yielding a reproducible scene graph. All geometric relation thresholds (e.g., maximum vertical gap, minimum footprint overlap, and horizontal proximity) are fixed across experiments and tuned empirically once per dataset.

The use of foundation models for perception, such as SAM and CLIP, is crucial in this context, as it removes the need for predefined categories, enabling recognition of unseen objects and better generalization to new environments.

\subsection{Active Semantic Perception}
To evaluate exploration efficacy, the framework incorporates two complementary strategies. As a geometric baseline, the Surface Edge Explorer (SEE) algorithm~\cite{border2024surface} uses measurement density to identify frontier points on partially observed surfaces and proposes viewpoints that maximize their visibility. In contrast, the Active Semantic Perception (ASP) framework~\cite{tang2025active} performs semantic-driven reasoning using the 3DSG produced by the RGB-only pipeline. At each step, an LLM samples plausible completions of the unobserved scene from the current graph. For each candidate viewpoint $x$ (among the navigable poses available in Habitat), ASP computes the expected information gain as the mutual information between the predicted observation $Y_{k+1}$ and the current graph $G_k$:
\begin{equation}
I(Y_{k+1}; G_k \mid x) = H(Y_{k+1} \mid x) - H(Y_{k+1} \mid x, G_k)
\end{equation}
where $H(\cdot)$ denotes Shannon entropy. By selecting the viewpoint that maximizes information gain, ASP prioritizes locations likely to resolve semantic ambiguities (e.g., behind a door leading to a kitchen). In this setting, the 3DSG acts not only as the target representation, but also as the cognitive structure guiding exploration. This dual role highlights the value of scene graphs, which compactly encode entities, attributes, and spatial relations while remaining expressive enough to support higher-level scene understanding and reasoning.
 
\subsection{Multi-View External Camera Integration}
Because the perception framework is strictly visual, observations from uncalibrated external RGB cameras can be processed by MapAnything identically to onboard egocentric images. This allows bird's-eye views to be incorporated into the same 3DSG pipeline to provide a broad initial estimate of the scene, update it during exploration, and improve scene context for downstream planning.

\section{Experiments and Results}
\label{sec:experiments}
We evaluate the method in three settings: static pipeline validation, dynamic active exploration, and multi-view external camera integration. Across all settings, node quality is assessed automatically using joint semantic and geometric matching: predicted and ground-truth labels are embedded with a SentenceTransformer~\cite{reimers2019sentencebert}, and matches are accepted only when both semantic similarity and 3D localization constraints satisfy fixed thresholds. Quantitative evaluation is therefore restricted to nodes, since datasets used in this work do not provide ground-truth relational annotations, while edges are generated by a deterministic geometry-based module that was extensively tested during development.

\subsection{Static RGB-Only 3DSG Pipeline Validation}
Baseline validation was conducted on the Replica dataset \cite{straub2019replica} to quantify accuracy lost in node prediction when substituting ground-truth depths and poses with MapAnything inferences. Table~\ref{tab:static_eval} reports the results.

\begin{table}[ht]
\centering
\caption{Quantitative evaluation of scene graph node accuracy across seven Replica scenes (\texttt{room0-2}, \texttt{office0-3}). CG is the original ConceptGraphs, CG-RGB is the proposed RGB-only variant, with predicted depths and poses. Values are reported as mean $\pm$ standard deviation across scenes.1}
\setlength{\tabcolsep}{5pt}
\begin{tabular}{lccc}
\toprule
Experiment & Precision $\uparrow$ & Recall $\uparrow$ & F1-score $\uparrow$ \\
\midrule
CG     & $\mathbf{0.686 \pm 0.05}$ & $0.401 \pm 0.10$        & $0.499 \pm 0.08$        \\
CG-RGB (ours) & $0.615 \pm 0.07$        & $\mathbf{0.436 \pm 0.12}$ & $\mathbf{0.500 \pm 0.08}$ \\
\bottomrule
\end{tabular}
\label{tab:static_eval}
\end{table}

The results demonstrate remarkable parity, with comparable F1 scores. MapAnything's noisier depth predictions introduce minor geometric inconsistencies, slightly lowering precision but preventing ConceptGraphs from overly aggressive merging of closely situated objects, thereby improving recall. Overall, these findings indicate that replacing ground-truth depth and pose with MapAnything estimates preserves the quality of scene graph nodes.

\subsection{Active Exploration Evaluation}

\begin{figure*}[t]
\centering
\begin{minipage}{0.48\textwidth}
    \centering
    \includegraphics[width=\linewidth]{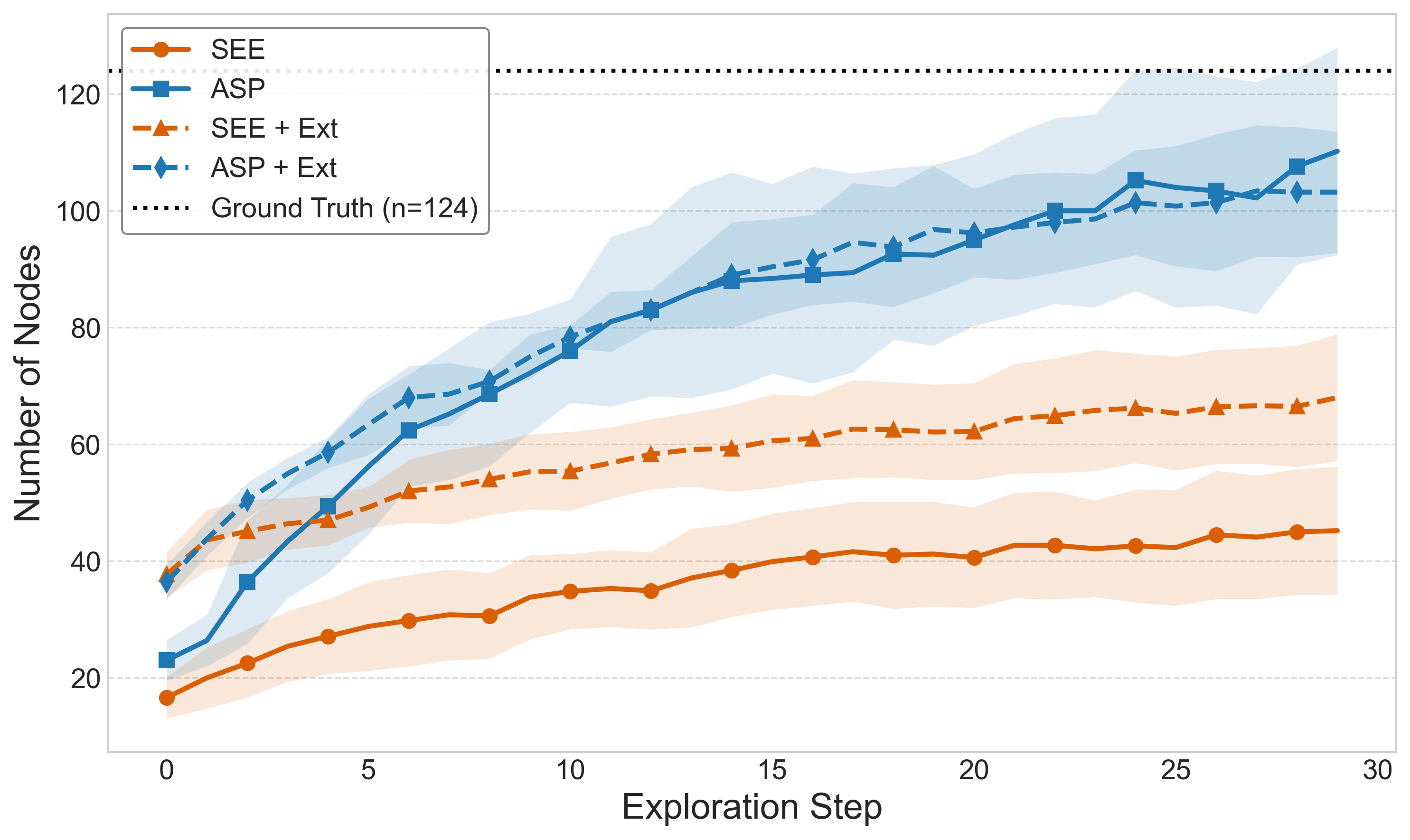}
\end{minipage}
\hfill
\begin{minipage}{0.48\textwidth}
    \centering
    \includegraphics[width=\linewidth]{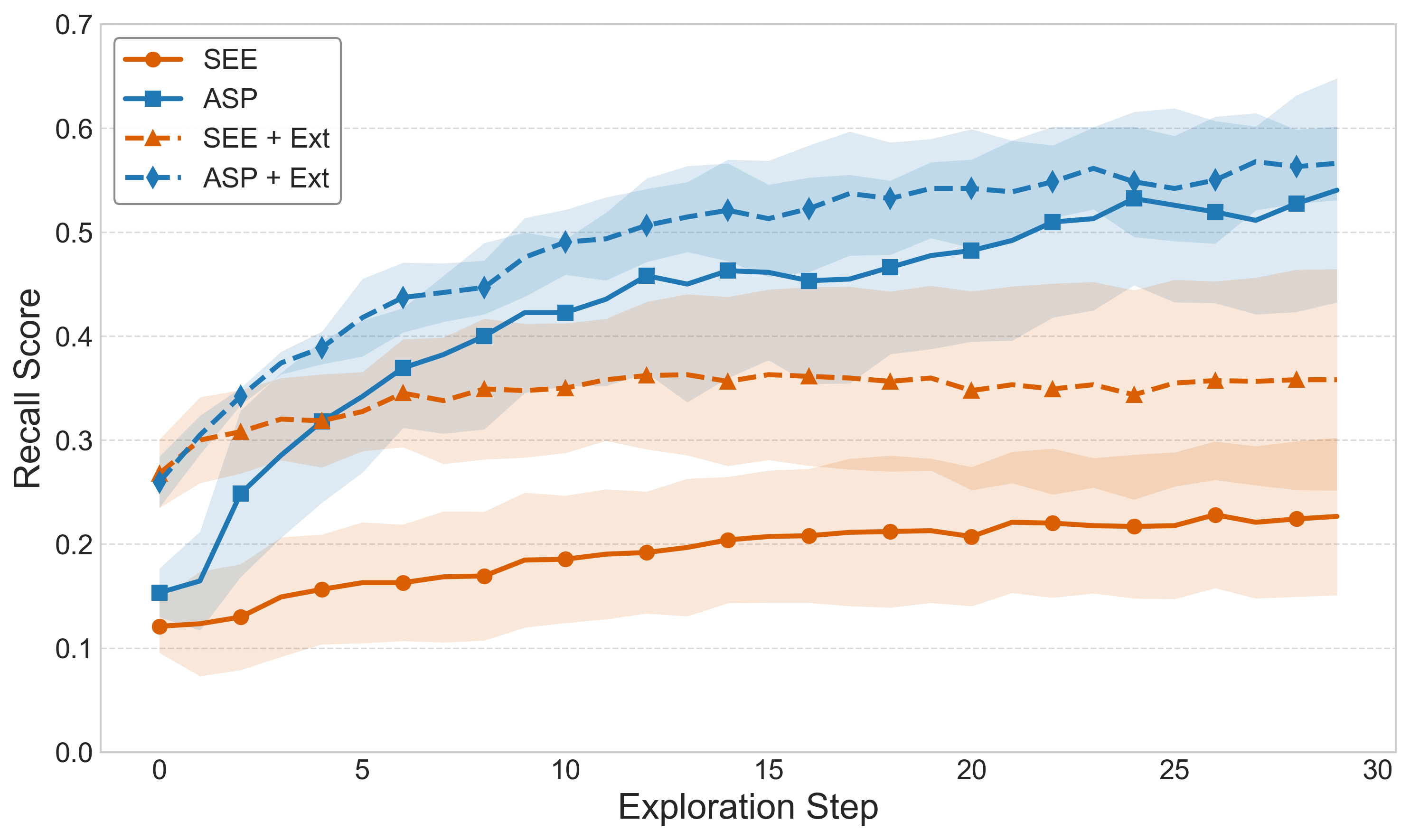}
\end{minipage}
\caption{Evolution of number of predicted nodes (left) and recall (right) for the geometry‑based (SEE) and semantic‑based (ASP) exploration algorithms, with and without an initial external camera view (+Ext), over exploration steps. Shaded areas denote standard deviation over different initial robot poses.}
\label{fig:active_expl}
\end{figure*}

The RGB-only active exploration pipeline reported in Fig.~\ref{fig:overview} was evaluated within the Habitat simulator \cite{savva2019habitat} using the six \texttt{apartment} scenes from the ReplicaCAD dataset~\cite{szot2021habitat2}, where local navigation to selected viewpoints is handled via Habitat’s built-in navigation utilities. \texttt{gemini-2.5-pro}~\cite{team2023gemini} was used as LLM for ASP. Experiments were run from 10 distinct starting positions over 30 exploration steps. Since performance trends were consistent across the six apartments, the results reported in Fig.~\ref{fig:active_expl} focus on \texttt{apartment 3} as a representative scene.

The semantic-driven ASP framework consistently outperformed the geometric SEE baseline. Operating exclusively on onboard cameras, ASP identified approximately 110 object nodes by step 30, approaching the ground-truth total (124). SEE stalled at roughly 45 nodes. In addition, ASP achieved double the recall (0.54 vs. 0.22). By prioritizing semantic ambiguity, ASP matches the quality of hundreds of passive trajectory frames in thirty intelligent movements.

\subsection{Multi-View External Camera Integration Study}
The efficiency of multi-view perception was tested by integrating fixed external infrastructure as bootstrapping inputs. In the active exploration trials (Fig.~\ref{fig:active_expl}), initializing the scene graph with just \textit{one} overhead external camera instantly jumped SEE's starting node count from 16 to 37 (+125\%) and ASP's from 23 to 36 (+57\%). Initial recall improved from 0.12 to 0.27 for SEE (+130\%) and from 0.15 to 0.27 for ASP (+80\%), and this was accompanied by a consistent improvement in final recall for both methods, particularly pronounced for SEE.

To evaluate standalone infrastructure capabilities, 270 experiments were conducted across 90 ReplicaCAD scenes utilizing only external cameras (from one to three) without a mobile robot. The scenes were split into complex \texttt{apartments}  (dense clutter, $\sim$123 objects, used also in the previous experiment) and simpler \texttt{furnished rooms} ($\sim$25 larger objects). Table~\ref{tab:external_camera_results} reports the results.

\begin{table}[ht]
\centering
\caption{Quantitative evaluation of scene graph node accuracy using between one and three fixed external cameras (\#Cam.) on ReplicaCAD \texttt{apartments} (Apt.) and \texttt{furnished rooms} (Fur.) scenes. Values are reported as mean $\pm$ standard deviation across scenes.}
\begin{tabular}{lcccc}
\toprule
Scene & \#Cam. & Precision $\uparrow$ & Recall $\uparrow$ & F1-score $\uparrow$ \\
\midrule
& 1 & \textbf{0.770 $\pm$ 0.054} & 0.198 $\pm$ 0.030 & 0.315 $\pm$ 0.041 \\
Apt. & 2 & 0.741 $\pm$ 0.036 & 0.232 $\pm$ 0.030 & 0.353 $\pm$ 0.036 \\
& 3 & 0.698 $\pm$ 0.031 & \textbf{0.301 $\pm$ 0.029} & \textbf{0.421 $\pm$ 0.030} \\
\midrule
& 1 & \textbf{0.566 $\pm$ 0.087} & 0.398 $\pm$ 0.071 & 0.465 $\pm$ 0.073 \\
Fur. & 2 & 0.540 $\pm$ 0.077 & 0.487 $\pm$ 0.071 & 0.510 $\pm$ 0.067 \\
& 3 & 0.486 $\pm$ 0.060 & \textbf{0.555 $\pm$ 0.065} & \textbf{0.516 $\pm$ 0.055} \\
\bottomrule
\end{tabular}
\label{tab:external_camera_results}
\end{table}

While 3 static viewpoints cannot match the recall of a 30-step active robotic exploration, they can successfully establish the environment's dominant structure. In complex apartments, three cameras achieved an F1-score of 0.421, while in furnished rooms, the F1-score reached 0.516, successfully recovering over 55\% of the ground truth objects instantly. Precision declines slightly as cameras are added due to overlapping multi-view duplicate fragments, but the net F1-score confirms that fixed external RGB cameras can provide useful complementary observations without requiring additional onboard sensing hardware.

\section{Discussion}
\label{sec:discussion}
The results suggest that active robot perception benefits more from structured semantic representations than from geometry alone. While dense geometric maps are effective for navigation, they do not explicitly encode object identity or spatial relations. By combining geometry, semantics, and structure, 3D scene graphs provide a richer and more informative perceptual model. Their structured nature also makes them naturally queryable by language models, which here act as spatial reasoning engines for inference in partially observed environments. Instead of relying only on geometric visibility, the system can use the evolving scene graph to reason about unobserved regions, infer plausible scene completions, and prioritize semantically informative viewpoints, consistent with the improved recall achieved by ASP.

Removing the dependence on depth sensors also offers practical advantages, since an RGB-only pipeline enables hardware-agnostic, multi-view perception and supports heterogeneous visual inputs. In this setting, fixed external cameras can help bootstrap and update the scene graph, reducing exploration effort while improving scene understanding. Beyond increasing scene graph completeness, these viewpoints can also support safer motion planning under semantic uncertainty by revealing obstacles, free space, and relevant structures before they become visible to the onboard camera, thereby reducing both geometric and semantic uncertainty in partially observed environments.

\section{Conclusion}
\label{sec:conclusion}
This paper introduced an active perception framework for incremental 3D scene graph construction using only RGB visual inputs.
By combining feed-forward reconstruction, open-vocabulary semantics, and LLM-driven exploration, the system removes depth-sensor dependency while improving exploration efficiency. Experiments demonstrate that semantic-driven perception significantly outperforms geometric baselines, while the RGB-only design enables effective integration of external cameras, opening broader possibilities for flexible, infrastructure-assisted perception in real-world robotic deployments. These results highlight the potential of semantics-first representations as a foundation for reliable and scalable robot autonomy.

\bibliographystyle{IEEEtran}
\bibliography{biblio}

\end{document}